# ReadMOF: Structure-Free Semantic Embeddings from Systematic MOF Nomenclature for Machine Learning

Kewei Zhu[1], Cameron Wilson [1], Bartosz Mazur, [1,2] Yi Li [1], Ashleigh M. Chester [1,3], and Peyman Z. Moghadam [1,3] *

1.  Department of Chemical Engineering, University College London, London WC1E 7JE, United Kingdom
2.  Department of Micro, Nano, and Bioprocess Engineering, Faculty of Chemistry, Wroclaw University of Science and Technology, Wroclaw 50-370, Poland
3.  Manufacturing Futures Lab, University College London, 7 Sidings Street, London, E20 2AE, United Kingdom
*  Corresponding author: Peyman Z. Moghadam (p.moghadam@ucl.ac.uk)

## Abstract

Systematic chemical names, such as IUPAC-style nomenclature for metal-organic frameworks (MOFs), contain rich structural and compositional information in a standardized textual format. Here we introduce ReadMOF, which is, to our knowledge, the first nomenclature-free machine learning framework that leverages these names to model structure-property relationships without requiring atomic coordinates or connectivity graphs. By employing pretrained language models, ReadMOF converts systematic MOF names from the Cambridge Structural Database (CSD) into vector embeddings that closely represent traditional structure-based descriptors. These embeddings enable applications in materials informatics, including property prediction, similarity retrieval, and clustering, with performance comparable to geometry-dependent methods. When combined with large language models, ReadMOF also establishes chemically meaningful reasoning ability with textual input only. Our results show that structured chemical language, interpreted through modern natural language processing techniques, can provide a scalable, interpretable, and geometry-independent alternative to conventional molecular representations. This approach opens new opportunities for language-driven discovery in materials science.

# 1    Introduction

Metal-organic frameworks (MOFs)[1, 2] are a class of crystalline porous materials constructed by the coordination of metal nodes, such as metal ions or metal-containing clusters, with multidentate organic linkers to form extended networks with internal porosity. Their modular and reticular nature enables precise control over pore size, geometry, chemical environment, and connectivity, which has led to widespread interest in applications such as energy storage[3], gas separation and purification[4], $CO_2$ capture[5], chemical catalysis[6, 7], environmental sensing[8], thermal regulation[9], and drug delivery[10]. The chemical and structural diversity of MOFs has positioned them as central materials in data-driven materials discovery. With tens of thousands of experimentally reported structures and millions of hypothetical frameworks now available through databases[11] and generative algorithms, machine learning and high-throughput computational screening have become indispensable tools for exploring the vast MOF design space[12]. For example, supervised learning models trained on descriptors derived from quantum mechanical calculations or molecular simulations have been applied to predict gas uptake capacities and mechanical or thermal stability[13]. These approaches significantly accelerate the identification of promising MOF candidates for targeted applications[14-17].

The structural complexity that enables the versatility of MOFs also presents significant challenges for their reliable computational characterization. Recent studies have shown that a non-negligible portion of structures in widely used computation-ready MOF databases[18-33], may contain fundamental chemical inconsistencies, including misassigned oxidation states for metal centers, missing hydrogens or coordinated solvent molecules[34, 35]. Since many predictive models rely heavily on accurate atomic coordinates and geometric descriptors, such structural inconsistencies can compromise predictive accuracy, lead to false positives, and undermine reproducibility. Moreover, experimental structures frequently suffer from disorder, missing atoms, or ambiguous unit cells, and even minor atomic perturbations can cause significant deviations in computed properties. These limitations highlight the fragility and preprocessing burden of conventional structure-based representations, specifically those relying on atomic geometries and crystallographic coordinates, which are highly sensitive to imperfections in experimental data.

Considering these limitations, alternative representations that remain chemically expressive yet less dependent on complete structural information would be valuable. One such alternative lies in systematic chemical nomenclature. The International Union of Pure and Applied Chemistry (IUPAC)-style[36] names for MOFs encode critical information, including metal identity, ligand composition, connectivity, coordination environment, and dimensionality. These chemically grounded descriptors remain accessible even when atomic structures are incomplete, ambiguous, or noisy. As demonstrated by recent efforts to curate reliable subsets from crystallographic databases[20], obtaining high-fidelity MOF structures remains nontrivial. Thus, systematic names present a robust and interpretable representation format that can serve as a complementary or even primary input for classification, similarity analysis, or machine learning, particularly in scenarios where structure-based approaches are unreliable. For example, in the nomenclature of MOFs, the prefix "*catena-*" is commonly used to denote that the structure exhibits infinite extension in one or more dimensions, a characteristic feature commonly associated with MOFs. Compound names following this convention are typically enclosed in brackets and provide a detailed description of the repeating structural unit. These names include the types of bridging ligands, indicated by the Greek letter **μ**, such as *μ-, μ2, μ3-*, the specific chemical identity of the ligands, and the metal centers, often with their oxidation states specified. Complex organic ligands are named according to IUPAC rules, and metal nodes are usually described by their nuclearity and oxidation state (e.g., *di-copper(I)*, *tri-nickel(II)*). **Figure 1** shows the systematic name of a common MOF, IRMOF-1[37]: *catena***-(tris(μ4-terephthalato)-(μ4-oxo)-tetra-zinc)**, encapsulating its essential chemical features. Specifically, the term **tetra-zinc** refers to a $Zn_4O$ (tetra-zinc) cluster, wherein four **zinc** ions are bridged by a central **μ4-oxo** ligand. Each **μ4-terephthalato** ligand binds four metal centers, reinforcing the extended connectivity. Despite their richness in structural information, such IUPAC-based names have been largely overlooked as direct input features for computational modeling and representation learning of MOFs. They capture chemically meaningful details of molecular building blocks, coordination geometry and network architecture, and can serve as a latent, underutilized source of features for machine learning-driven discovery.

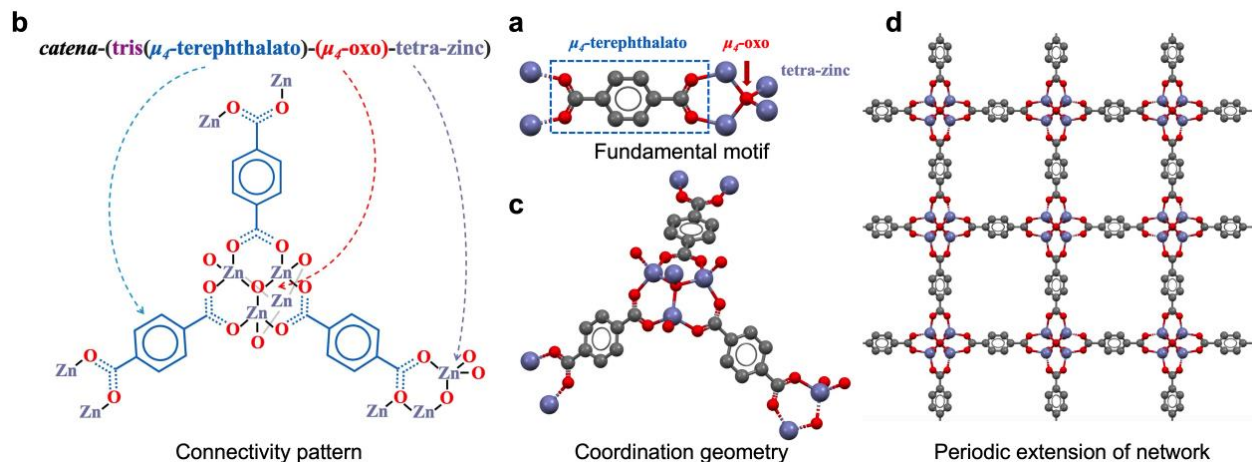

**Figure 1. Structural representation of IRMOF-1[37] derived from its IUPAC-style name,** *catena***-(tris(μ4-terephthalato)-(μ4-oxo)-tetra-zinc).** (a) The fundamental secondary building unit (SBU) comprises a $Zn_4O$ cluster coordinated by **μ4-terephthalato** and **μ4-oxo** ligands. (b) The connectivity pattern reflects the threefold bridging of terephthalate linkers to the central $Zn_4O$ core. (c) This arrangement gives rise to a rigid, symmetric coordination geometry at the node level. (d) Periodic propagation of the motif in three dimensions forms the extended crystalline framework.

Recent advances in natural language processing (NLP), particularly transformer-based language models, have enabled the extraction of rich semantic structure from complex textual data[38, 39]. These model are trained using techniques such as contrastive learning[40, 41], which encourages discrimination between similar and dissimilar examples; multi-task training[42, 43], which promotes generalizable representations across related objectives; and instruction tuning[44, 45], where models are refined using natural language prompts to better align with specific tasks. Such strategies have led to impressive cross-domain generalization, extending NLP models into scientific domains such as chemistry[46, 47] and biology[48, 49]. For example, Ilharco *et al.*[50] showed that large language models can perform image classification through textual abstraction, while AlphaFold[51] demonstrated that deep sequence embeddings can predict 3D protein structures directly from amino acid sequences without explicit structural input. These developments collectively suggest that text-based sequence representations can act as effective surrogates for structural modeling, particularly when the input encodes chemically or biologically meaningful patterns. In the field of chemistry, prior works such as Mol2Vec[52] and Chemformer[53] have leveraged SMILES strings or molecular graphs to learn molecular representations. However, these methods primarily target small molecules and do not incorporate systematic IUPAC-style nomenclature, especially for reticular materials like MOFs. In contrast, our work treats the formal and standardized IUPAC-like name itself as a chemically expressive representation, bypassing the need for atomic coordinates or connectivity graphs, and thereby differs from prior coordinate-free approaches, such as RFcode[54], MOFid/MOFkey[55], MOFormer[56] and MOFGPT[57] and stoichiometry-based descriptors[58], that rely on engineered identifiers or structure-derived strings rather than systematic chemical nomenclature. This approach offers a novel pathway for modeling structure-property relationships in porous materials by relying solely on structured chemical language, thereby addressing a key gap left by existing NLP-based molecular encoders.

To operationalize this concept, we introduce **ReadMOF**, a structure-free method that derives chemically meaningful representations from systematic MOF names using pretrained language models. Instead of relying on 3D coordinates or structural graphs, ReadMOF tokenizes and encodes IUPAC-like names using a pretrained language model. The generated name embeddings, which are vector representations learned from text, capture latent semantic patterns that reflect underlying chemical composition, including metal identity and ligand class. **Figure 2a** illustrates how our method captures chemically meaningful patterns from systematic MOF names. Each name is tokenized and encoded by a pretrained language model, yielding high-dimensional vector representations. These high-dimensional embeddings are then projected into two dimensions for visualization. As shown in the right panel of **Figure 2a**, complexes with identical ligands but different metal centers (e.g., cobalt *versus* nickel) exhibit systematic spatial displacements in the embedding space. This observation suggests that the model can recognize chemical substitutions directly from textual input. Specifically, each chemical token, such as "cobalt," "nickel," or "quinoline," is assigned to a context-dependent embedding based on its co-occurrence with other terms in the name. As a result, metal-ligand environments with similar chemical features tend to cluster together in the embedding space. Notably, metal substitutions produce consistent directional shifts in embedding space regardless of the surrounding ligand. For example, replacing cobalt with nickel yields nearly parallel vector displacements across different ligands. From a coordination chemistry perspective, this behavior reflects an emergent understanding of periodic and electronic similarity among transition metals. By organizing nomenclature-derived data in a structured vector space, ReadMOF enables predictive modeling and systematic analysis of compositional variation without requiring explicit atomic structures.

An overview of the ReadMOF pipeline is shown in **Figure 2b**. The method begins by encoding systematic MOF names using pretrained language models to generate continuous vector embeddings, which serve as chemically meaningful representations derived entirely from text. We first evaluate these name embeddings by comparing them to traditional structure-based descriptors, specifically revised autocorrelation descriptors (RACs)[59]. Using cosine similarity and retrieval tasks, we assess whether the text-derived embeddings preserve structural consistency, such as grouping similar metal-ligand environments or framework types. The comparison reveals that name embeddings capture comparable compositional trends, despite lacking explicit atomic information. To further assess interpretability, we apply dimensionality reduction techniques to project the high-dimensional embeddings into two dimensions. The resulting maps reveal chemically coherent clusters and consistent vector shifts corresponding to metal substitution or ligand variation, indicating that the learned space encodes latent structure-property relationships. Beyond structural comparisons, the embeddings support downstream applications such as property prediction (e.g., bandgap, pore volume) and generative reasoning tasks. Importantly, name-based predictions can be connected to experimentally reported conductive or semiconductive MOFs and extended to the identification of candidate materials for experimental validation. Additionally, large language models conditioned on name embeddings can infer chemical formulas or propose plausible synthesis strategies. Importantly, the entire pipeline operates without requiring atomistic models or crystallographic data. This makes ReadMOF particularly suited for high-throughput screening, early-stage materials discovery, or scenarios where structural information is incomplete or uncertain. Compared to graph-based molecular representations, which often require extensive preprocessing and are sensitive to structural noise, our approach offers a scalable and interpretable alternative rooted in chemical language.

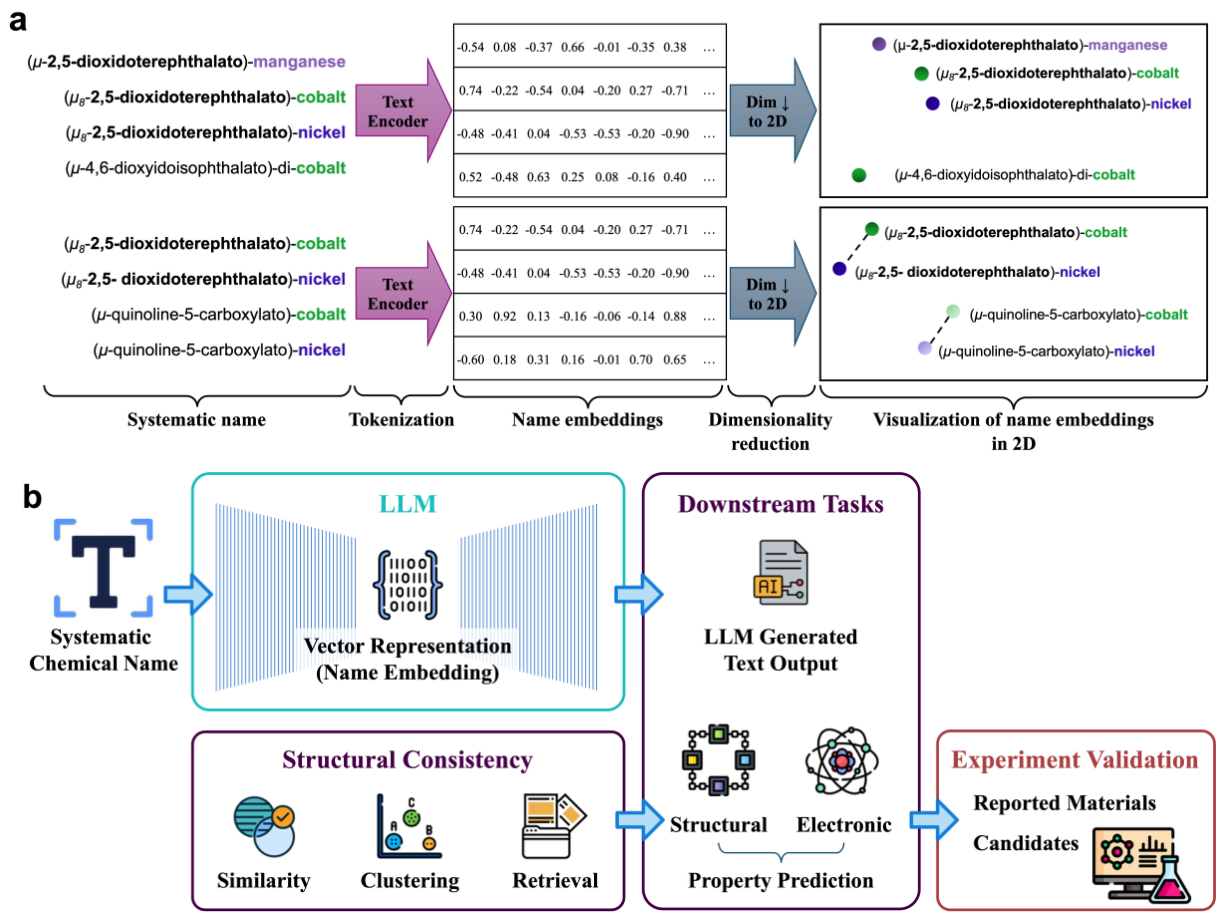

**Figure 2.** (a) Text-derived embeddings capture chemically meaningful relationships in MOF nomenclature. Systematic chemical names are tokenized and encoded using a pretrained language model, yielding continuous embeddings that preserve metal-ligand similarity and systematic shifts due to metal substitution (e.g., cobalt ↔ nickel). (b) Schematic illustration of the ReadMOF pipeline for language-based representation learning. Systematic names are transformed into embeddings, which support structural tasks (similarity, retrieval, clustering) and enable downstream applications including structural and electronic property prediction as well as generative reasoning, without requiring atomic structures or crystallographic coordinates as input. Furthermore, these embedding-derived predictions can be mapped to experimentally reported materials to prioritize candidates for future laboratory validation.

This work demonstrates that systematic chemical names, when processed using pretrained language models, can serve as chemically grounded and scalable representations for molecular machine learning. By leveraging the implicit chemical logic encoded in standardized nomenclature, our approach enables structure-free modeling workflows that are interpretable, data-efficient, and broadly applicable. We show that this method performs effectively across tasks such as structural retrieval, clustering, and property prediction for both real and hypothetical MOFs. Furthermore, the underlying principle is readily extensible to other classes of reticular materials, offering a promising direction for scalable, language-based representations in materials informatics. For clarity, a detailed analysis of the information completeness and ambiguity boundaries of systematic nomenclature, including a comparison between geometry-based and chemically abstracted representations, is provided in Supplementary Information Sections S1.1.

## 2   Results and discussion

To determine whether systematic IUPAC-style nomenclature can serve as a structure-independent yet chemically informative representation for metal–organic frameworks (MOFs), we evaluated name-based embeddings across multiple levels of analysis, spanning semantic similarity, unsupervised organization, retrieval, and electronic property prediction. Using filtered subsets of the Cambridge Structural Database (CSD)[20, 60, 61], we find that embeddings derived solely from systematic names encode chemically meaningful relationships that closely mirror trends obtained from structure-based descriptors, despite the complete absence of explicit geometric input.

Importantly, extending this framework to electronic property prediction allows direct connection to experimentally relevant behavior: bandgap predictions based only on systematic names significantly enrich electronically active regions of MOF chemical space, including materials previously reported to exhibit semiconductive or conductive behavior. Leveraging these predictions, we further identify and propose a set of chemically plausible MOF candidates with low predicted bandgaps as promising targets for experimental investigation.

Finally, we discuss how systematic chemical nomenclature naturally interfaces with large language models, enabling generative and reasoning tasks that extend beyond property prediction toward data-driven MOF discovery.

## 2.1    Semantic alignment with structure-derived representations (RAC)

In this section, we investigate whether embeddings derived from systematic MOF nomenclature can reproduce chemically meaningful similarity relationships captured by structure-derived descriptors. We first establish a quantitative benchmark by comparing semantic similarity derived from MOF names with structure-based similarity computed from revised autocorrelations (RACs)[62], following the methodology of Moosavi *et al*[59], thereby enabling a quantitative comparison across different text encoders. RACs encode local chemical environments through atom-centered autocorrelations of atomic properties along bonds and paths and are widely used in MOF structure-property modeling. Calculation details are provided in Supporting Information S1.2.

All analyses were performed on a filtered subset of 31,103 polymeric MOFs from the CSD with validated systematic names (Supplementary Information S7), providing a high-quality basis for semantic similarity and alignment analysis. Using the Filtered Set, we evaluated 27 publicly pretrained text encoders from the ChemTEB[63] benchmark in a zero-shot setting, including representative models from families such as SBERT[64, 65], BGE[66, 67], E5[68] and Nomic AI[69], ChemicalBert[70, 71], MatSciBERT[72], SciBert[73], BERT[70]. All models were used as released, with their native tokenization schemes, and without any additional chemistry-specific input preprocessing or feature engineering. Among these models, *nomic-embed-v1.5 encoder* exhibited the highest consistency and is used as a representative case in the main text; complete benchmarking results are summarized in Table S1 of the Supporting Information.

We examined semantic alignment between MOF name embeddings and structure-derived chemical similarity by performing a second-order similarity analysis. Specifically, a cosine similarity matrix of name embeddings was computed across the MOF set and compared it with a RAC-based similarity matrix. The cosine similarity between these matrices quantified the alignment between semantic similarity (from names) and chemical similarity derived from structure-based descriptors. This procedure was applied to all 27 text encoders with full visualizations provided in Supporting Information Figure S2-1, and the corresponding numerical values are summarized in the Cos. Similarity column of Supporting Information Table S1.

Among all models, the *nomic-embed-v1.5 encoder* exhibited the strongest semantic-structural consistency, with a cosine similarity of 0.96, and is therefore used as the representative case in the main text. **Figure 3a** presents the heatmap of the cosine similarity matrix for MOF name embeddings generated by this encoder. Distinct diagonal and block structures indicate that the embeddings preserve chemically meaningful similarity relationships driven by composition and local chemical environment, consistent with trends captured by structure-derived descriptors. These results show that language models can capture chemically relevant similarity patterns directly from systematic nomenclature, without explicit geometric input.

Having identified a representative encoder (*nomic-embed-v1.5*), we visualize the global organization of the name-embedding space. To provide qualitative insights, MOF name embeddings were projected into two dimensions using t-SNE. **Figure 3b** shows that the embedding space exhibits well-defined clusters dominated by transition metals such as Cu, Co, Ni, and Zn. Notably, these distributions were obtained solely from systematic names, without the use of atomic coordinates or geometric descriptors, indicating that dominant compositional factors are encoded directly in nomenclature text. These results demonstrate that systematic MOF names induce a chemically organized latent space that supports chemically informed interpretation of compositional trends. Complete visualizations for all 27 encoders are provided in Supporting Information Figure S4-1.

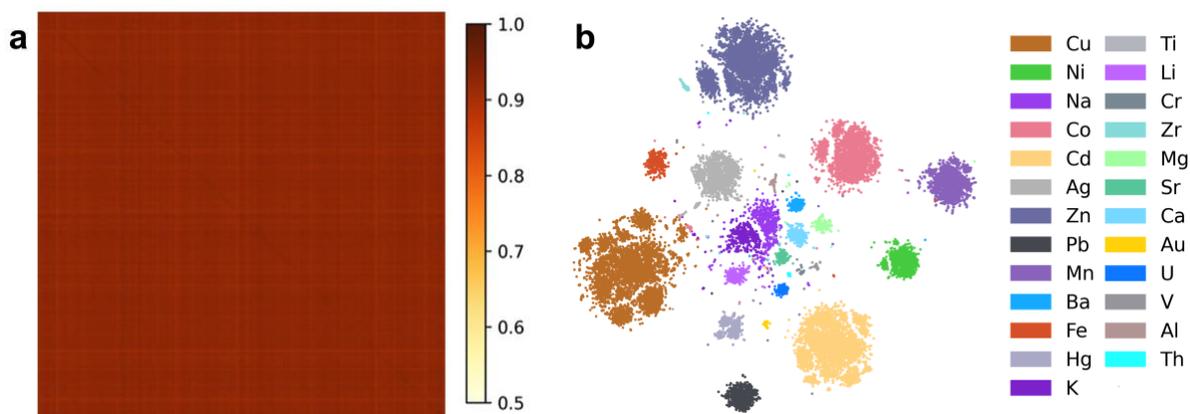

**Figure 3.** Semantic organization of MOF names using the *nomic-embed-v1.5* encoder. (a) Cosine similarity matrix of name embeddings across the MOF set, showing strong diagonal and block structures that indicate chemically meaningful groupings. **(b)** Two-dimensional t-SNE projection of the same embeddings, colored by metal identity. Distinct clusters corresponding to metals (e.g., Cu, Co, Ni, Zn) highlight that systematic nomenclature encodes compositional information that can be effectively captured by semantic models.

## 2.2 Chemical information encoded in systematic MOF nomenclature

To provide chemical interpretation of the observed embedding organization without invoking explicit structural reconstruction, we examine how bridging modes ($\mu$), as specified in systematic nomenclature, contribute to chemically consistent grouping of MOFs in the embedding space. In this context, $\mu$ values are treated as descriptors of ligand coordination roles rather than as explicit representations of framework connectivity.

Systematic names provide a standardized description of secondary building units (SBUs) through coordination-role annotations, including the number of metal centers bridged by a ligand. **Figure 4** illustrates representative cases that highlight how such coordination-role descriptors give rise to chemically distinct but abstracted framework motifs. Panels (a-d) all involve terephthalate (BDC) ligands, yet differ in metal identity and coordination roles . For example, IRMOF-1[37] (**Figure 4a**) contains a $Zn_4O$ cluster described by $\mu_4$-O and $\mu_4$-carboxylates, while MOF-2[74] (**Figure 4b**) also described using $\mu_4$-BDC, adopts a chemically distinct layered motif . UiO-66[75] (**Figure 4c**) employs $Zr_6$-based nodes with high coordination multiplicity, whereas MIL-88B[76] (**Figure 4d**) features $Fe_3O$-based coordination environments associated with flexible framework behavior. Beyond terephthalate systems, panels (e-g) further illustrate how coordination-role descriptors differentiate chemically distinct framework motifs. ZIF-8[77] (**Figure 4e**) employs $\mu_2$-2-methylimidazolate ligands associated with $ZnN_4$ tetrahedral nodes, while HKUST-1[78] (**Figure 4f**) and DUT-10[79] (**Figure 4g**) involve higher coordination multiplicities that correspond to chemically distinct node motifs. These examples are not intended to imply recovery of detailed connectivity, but rather to illustrate that systematic nomenclature encodes chemically meaningful coordination roles that are consistently reflected in the embedding space.

The distribution of ligand coordination roles within the embedding space is summarized in **Figure 4h**. Ligands annotated with higher coordination multiplicities ($\mu_4$–$\mu_8$) form relatively compact clusters, reflecting the consistency of their coordination roles across chemically related frameworks, whereas lower-coordination descriptors appear more dispersed. These results demonstrate that systematic MOF nomenclature induces chemically constrained signals in the embedding space, supporting interpretation at the level of chemical roles rather than explicit connectivity.

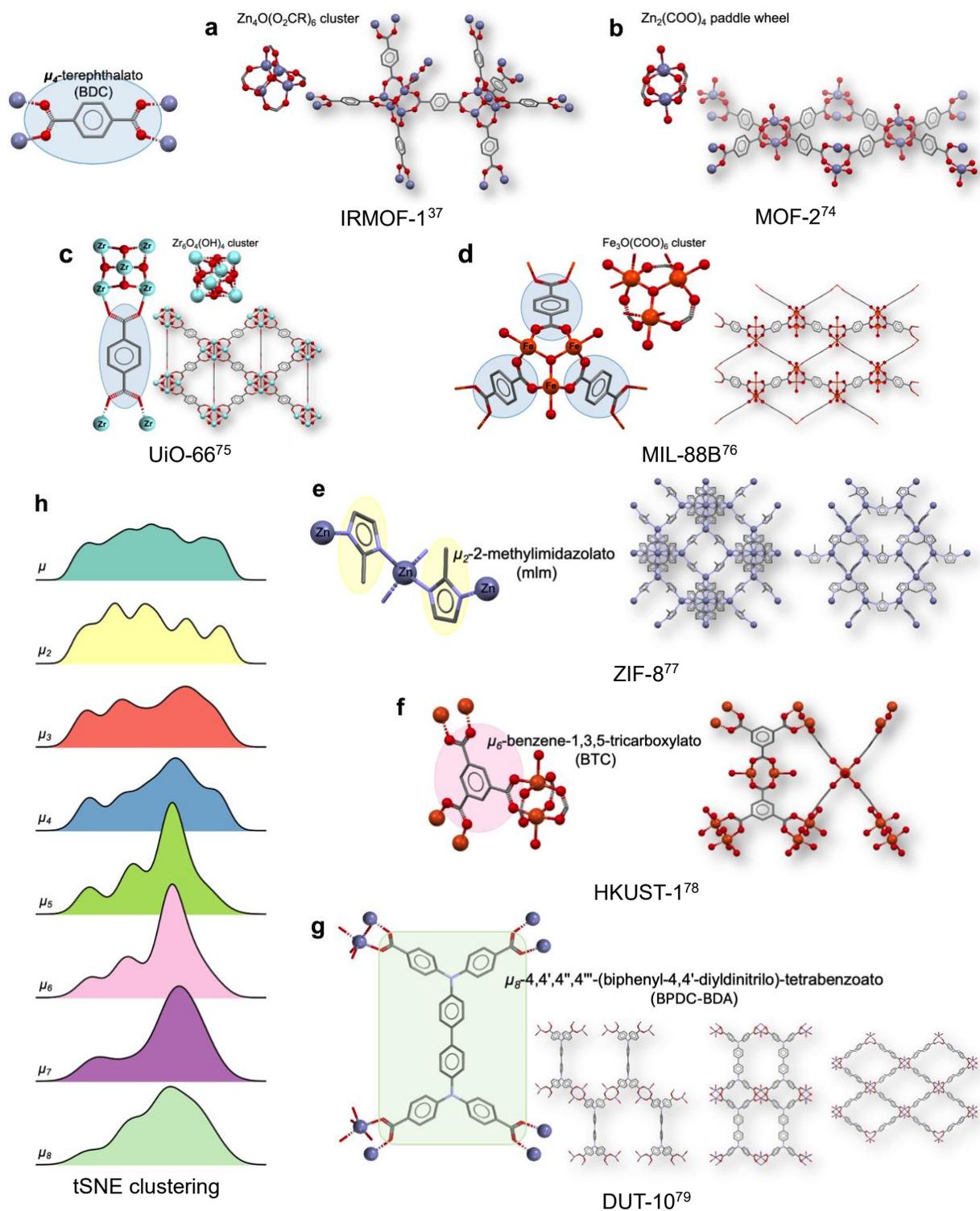

**Figure 4. Representative MOF structures and ligand connectivity patterns revealed through systematic chemical names. (a-d)** Terephthalate (BDC)-based frameworks illustrate how varying $\mu$ values and metal nodes yield distinct coordination features, from robust 3D $Zn_4O$ clusters

(IRMOF-1) to flexible Fe₃O SBUs (MIL-88B). **(e–g)** Non-BDC systems highlight connectivity trends, from low-connectivity $\mu_2$-2-methylimidazolate (ZIF-8) to highly crosslinked $\mu_6$-BTC (HKUST-1) and $\mu_8$-BPDC-BDA (DUT-10). **(h)** The distribution of $\mu$ coordination environments in the embedding space. Structures are reproduced with permission from ©CCDC.

## 2.2    Semantic retrieval reflects chemical abstraction level

We further evaluate the chemical abstraction encoded in systematic nomenclature supports semantically faithful retrieval across variations in composition and naming conventions. To assess semantic fidelity, we performed a retrieval task in which MOFs were ranked by cosine similarity of name embeddings and evaluated for shared features such as metal identity and ligand class. Retrieval behavior based on name-based embeddings was compared with that obtained using RAC-based descriptors, which provide a structure-derived chemical similarity reference. Summary results for all encoders and retrieval metrics are reported in Supporting Information Table S1, and a representative example is shown in **Figure 5** with associated reference properties listed Table 1.

The retrieval target in **Figure 5a** is a cobalt terephthalate. Using a fixed similarity cutoff (cosine similarity ≥ 0.85), Structure 5b is retrieved by both RAC and Name embeddings and differs from the target only by metal substitution (Co → Ni). Structure 5c, retrieved exclusively by RAC, reflects a manganese substitution and captures subtle geometric distinctions emphasized by structure-derived descriptors. In contrast, the name embeddings retrieve a broader semantic neighborhood: structures 5d, 5e, and 5f exceed the same threshold but are not identified by RAC-based retrieval. These results highlight that name embeddings prioritize chemical role similarity over strict geometric matching.

In particular, structure 5d differs from the target in metal identity (Co → Fe) and naming convention ("terephthalato" vs. "1,4-benzenedicarboxylato") yet is retrieved as chemically related.

Structures 5e and 5f are retrieved despite variations in ligand isomerism and crystallographic symmetry, indicating that name embeddings capture chemically relevant descriptors preserved in systematic nomenclature even when detailed geometric features differ. Notably, the corresponding reference properties of these structures remain within approximately 5% of the target values, suggesting that semantic similarity aligns with overall structural comparability.

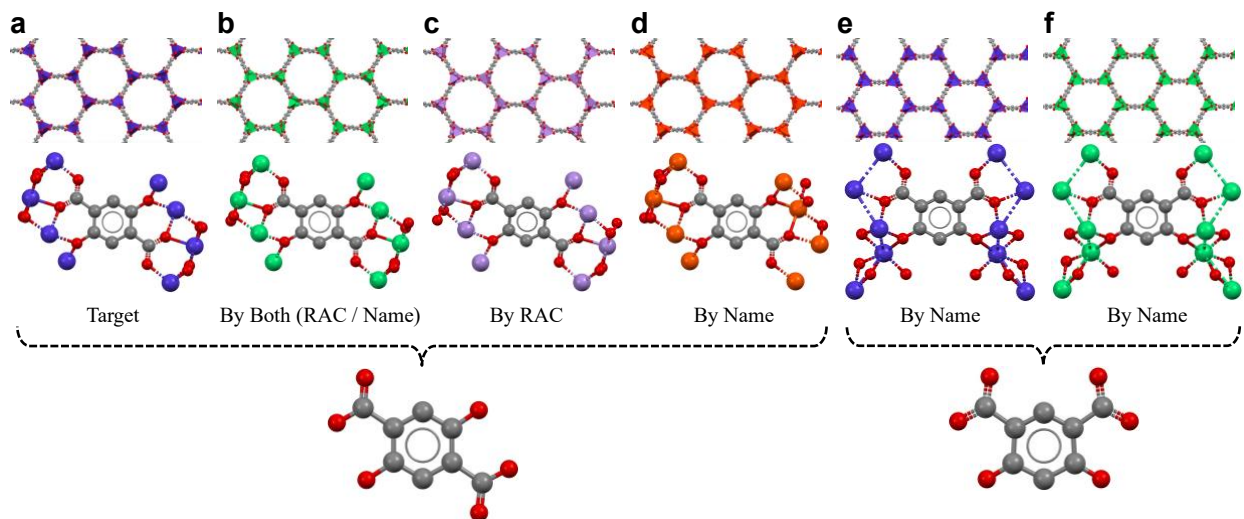

**Figure 5.** Retrieval performance for a target MOF **(a)** using *nomic-embed-v1.5* embeddings. Both RAC and name embeddings recover close analogs **(b)**, while RAC uniquely identifies subtle geometric variants **(c)**. Name embeddings, in contrast, retrieve additional structures with alternative metals, ligand variants, or naming conventions **(d–f)**, demonstrating their ability to generalize latent chemical cues beyond explicit structural descriptors. Structures are reproduced with permission from ©CCDC.

**Table 1.** Structural properties of MOFs retrieved using different embedding types.

| Embedding | Fig 5 | Chemical Name (*catena-…*) | Space Group | Formula | Cell Volume (Å³) | Density (g/cm³) |
|---|---|---|---|---|---|---|
| Retrieval Target | (a) | ($\mu_5$-2,5-dioxidoterephthalato)-cobalt(ii)[80] | R-3 | $(C_8H_2Co_2O_6)_n$ | 3962.09 | 1.117 |
| RAC / Name | (b) | ($\mu_8$-2,5-dioxidoterephthalato)-di-nickel(ii)[81] | R-3 | $(C_8H_2Ni_2O_6)_n$ | 3898.34 | 1.194 |
| RAC | (c) | ($\mu$-2,5-dioxido-1,4-benzenedicarboxylato)-di-manganese(ii)[82] | R-3 | $(C_8H_2Mn_2O_6)_n$ | 4222.73 | 1.076 |
| Name | (d) | ($\mu$-2,5-dioxido-1,4-benzenedicarboxylato)-di-iron[83] | R-3 | $(C_8H_2Fe_2O_6)_n$ | 4041.30 | 1.131 |
| Name | (e) | ($\mu$-4,6-dioxyidoisophthalato)-di-cobalt[84] | R3m | $(C_8H_2Co_2O_6)_n$ | 3923.55 | 1.118 |
| Name | (f) | ($\mu$-4,6-dioxyidoisophthalato)-di-nickel[84] | R3m | $(C_8H_2Ni_2O_6)_n$ | 3878.22 | 1.200 |

Taken together, these retrieval results demonstrate that name embeddings encode metal identity, ligand chemistry, and coordination-role information at an abstracted chemical level. By mapping chemically analogous but syntactically diverse frameworks to nearby regions of embedding space, they recognize semantic equivalence despite variations in naming, substitution, or geometry. The representation identifies semantic equivalence across variations in naming and minor structural differences. This behavior reflects the intended abstraction level of systematic nomenclature and makes name embeddings well suited for discovering chemically related framework variants rather than reproducing exact structural matches.

## 2.3 Emergent property prediction from systematic IUPAC chemical names

Next, we assessed whether name embeddings alone are sufficient for downstream property prediction, treating predictive performance as a test of whether the chemical abstraction encoded in systematic nomenclature is adequate for quantitative modeling. We hypothesized that systematic MOF names, while lacking explicit structural coordinates, nonetheless encode sufficient chemical information to support predictive modeling of both geometric and electronic descriptors.

Using filtered CSD-derived datasets, supervised regressors trained on name embeddings achieved strong performance in predicting both structural and electronic properties of MOFs. Specifically, models trained on the CSD-MOF set from the CSD[61] accurately predicted porosity-related metrics and other geometric descriptors, while embeddings derived from the CSD-QMOF set (10,716 entries from CSD[61] and QMOF[21, 22] intersection) enabled prediction of DFT-computed bandgaps across multiple exchange-correlation functionals. All models were trained using 10-fold cross-validation, and ablation studies confirmed that distinct components of systematic nomenclature contribute critically to predictive accuracy.

**Figure 6** provides an overview of these results, with panel 6a illustrating ligand-dependent structural properties and panel 6b highlighting metal-dependent electronic properties. For completeness, a quantitative performance comparison with representative structure-driven models under identical data and evaluation settings is reported in Supporting Information Section S5.3, with the corresponding complexity analysis presented in Section S5.4.

For **structural properties**, models trained on name embeddings achieved consistently high predictive accuracy, with $R^2$ values above 0.88 for largest cavity diameter (LCD), accessible surface area (ASA), density, and void fraction (Supporting Information Figure S5-1). As shown in **Figure 6**a, the predicted distributions closely match simulation-derived reference values for MOFs containing three widely used linkers: terephthalic acid (BDC), trimesic acid (BTC), and 4,4′-bipyridyl. These results indicate that name embeddings reflect ligand- dependent chemical constraints that systematically influence framework geometry. For example, BTC-based MOFs exhibit broader range of geometric properties than BDC-containing frameworks, consistent with the additional branching introduced by the tricarboxylate motif, while 4,4′-bipyridyl tend to show narrower pores and higher densities. Ablation analyses further confirmed the central role of organic linkers and their coordination environments: removing ligand-related terms caused the most significant performance degradation, whereas masking structural modifiers had comparatively little effect (Supporting Information Figure S5-2), consistent with the abstracted nature of the representation.

For **electronic properties**, models trained solely on systematic names also achieved strong performance, with $R^2$ exceeding 0.90 across multiple DFT functionals (Supporting Information Figure S5-5). **Figure 6b** presents the predicted PBE bandgap distributions stratified by metal identity and oxidation state. Open-shell cations such as $Cu^{2+}$, $Ni^{2+}$, and $Fe^{2+}$ yield broader distributions centered at lower bandgap values, reflecting the influence of partially filled d orbitals, whereas closed-shell cations including $Zn^{2+}$, $Cd^{2+}$ and $Mg^{2+}$ produce narrower, higher-bandgap distributions. These patterns demonstrate that electronic trends are governed primarily by chemical identity rather than precise geometric detail. Consistent with this interpretation, ablation analyses show that masking metal-related terms led to the most severe performance degradation, while ligand removal caused a moderate decline and $\mu_x$ indicators contributed relatively little (see Supporting Information Figure S5-7 for details).

Together, these results demonstrate that embeddings derived from systematic nomenclature not only recover established chemical trends but also clarify which components are most influential for predicting structural versus electronic properties.

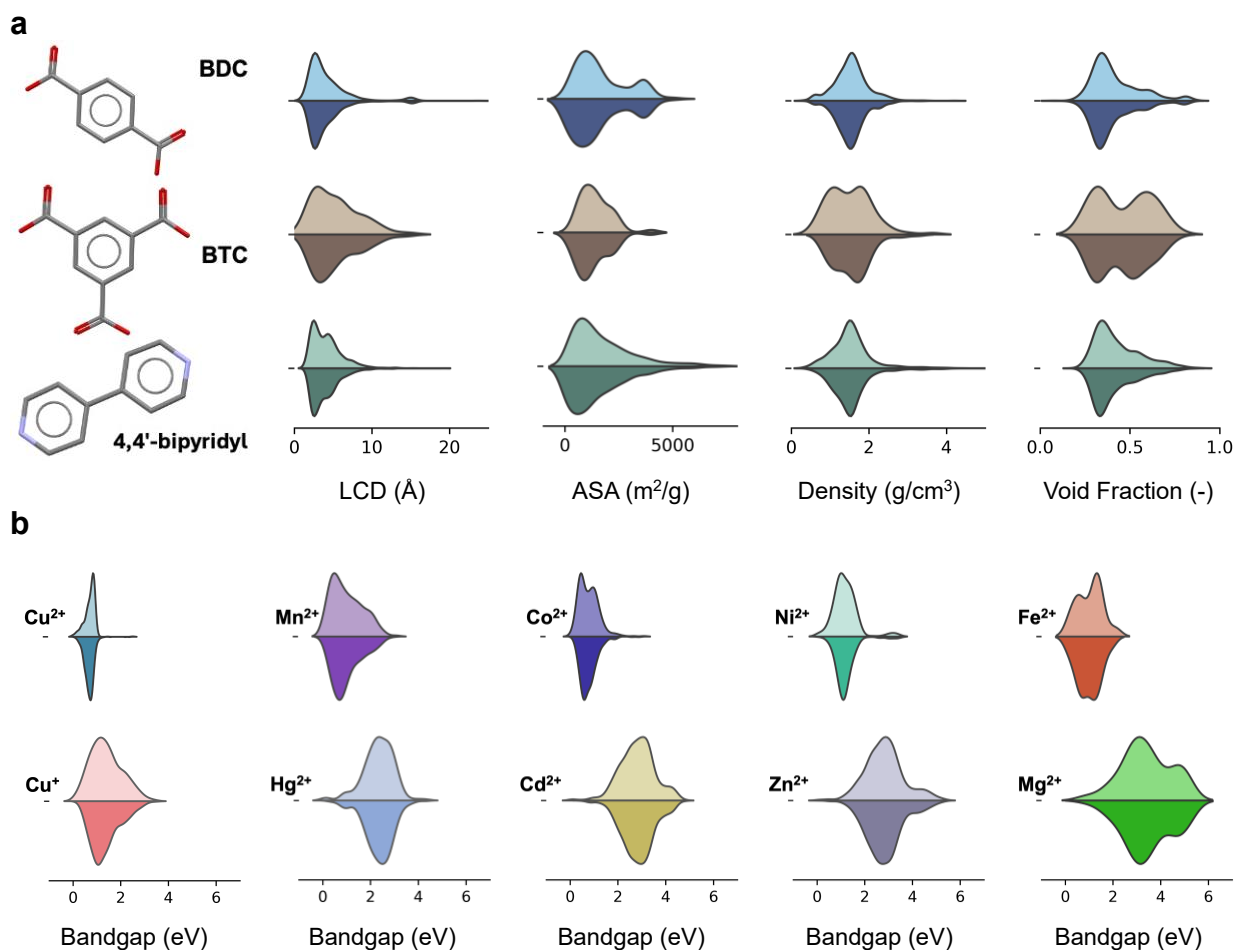

**Figure 6.** Comparative evaluation of structure-free name embeddings in predicting MOF properties. **(a)** Predicted (lower violins) and ground truth (upper violins) distributions of four structural properties (largest cavity diameter, LCD; accessible surface area, ASA; density; and void fraction) for MOFs containing three representative linkers. The agreement between predicted and reference distributions demonstrates that name embeddings capture ligand-dependent trends in framework geometry and porosity. **(b)** Predicted PBE bandgap distributions stratified by metal identity and oxidation state. Open-shell cations (e.g., $Co^{2+}$, $Ni^{2+}$, and $Fe^{2+}$) exhibit broader low-gap distributions due to partially filled d orbitals, whereas closed-shell cations (e.g., $Zn^{2+}$, $Mg^{2+}$, and $Cd^{2+}$) show narrower, higher-energy distributions consistent with electronic localization. These findings highlight that systematic nomenclature embeddings can recover chemically meaningful variability in both structural and electronic properties without explicit structural input.

## 2.4  Name-based screening for conductive MOFs

To demonstrate a practical deployment scenario of name-based representations, the bandgap models detailed in Section 2.3 were deployed as a front-end screening tool for 105,328 previously unseen structures from the Cambridge Structural Database (CSD) MOF subset. To ensure rigorous validation, all entries overlapping with the QMOF dataset were explicitly excluded. Given that electric conductivity in MOFs is strongly correlated with narrow bandgaps, these models serve as an efficient proxy for identifying conductive frameworks. As systematic nomenclature relies solely on textual metadata rather than fully resolved atomic coordinates, it represents the most accessible information during early-stage database exploration. In this context, the name-based model functions as a rapid, chemically informed prioritization tool for identifying electrically conductive MOFs. Figure 7 visualizes the predicted HSE06-PBE bandgap landscape for all CSD structures. The top 50 CSD MOFs with the lowest-bandgap predictions (HSE06 bandgap $\leq 1.5$ eV) are listed in Table S4 in the Supplementary Information. This ranking was benchmarked to determine the model's capacity to re-discover experimentally validated conductive MOFs and to identify new candidates.  The corresponding PBE predictions are included to illustrate the systematic relationship between the two functionals, with PBE consistently yielding lower bandgaps, in agreement with prior studies[85]. Among the top 50 structures, 18 structures that have been previously reported as semiconducting or conductive (Table 2) are highlighted as larger yellow-to-green circles, colored by their experimentally reported conductivity. The remaining candidates, for which no conductive or semiconductive behavior has been reported, are denoted as blue crosses. The substantial overlap between the top-ranked re-discovered candidates and experimentally validated conductive MOFs (18 out of 50) demonstrates a high screening precision underscoring the effectiveness of name-based representations as a computationally efficient filter for identifying electrically active frameworks. The unexplored candidates further represent high-priority targets for future experimental characterization or high-level computational studies.

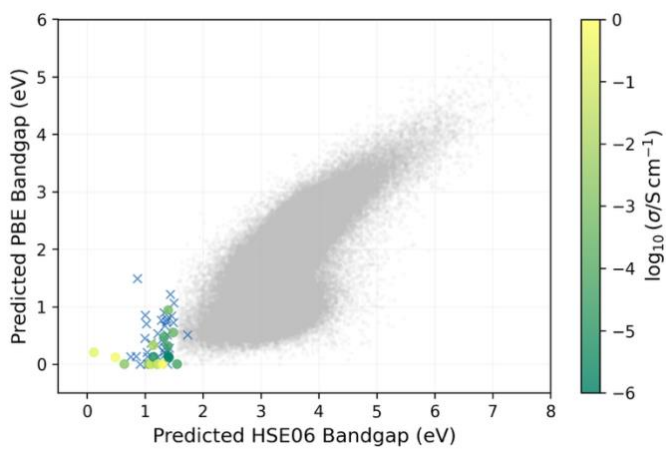

**Figure 7.** Predicted HSE06-PBE bandgap landscape of MOFs obtained from name-based models. Silver background scatter points denote the full set of CSD MOF bandgap predictions. Already-known conductive structures are color coded by their reported electrical conductivity values. Blue crosses denote structures candidates for which no conductive or semiconductive behavior has been previously reported.

**Table 2.** 18 experimentally validated conductive and semiconductive MOFs (conductivity $> 10^{-6}$ S cm$^{-1}$) identified within the top 50 candidates ranked by predicted HSE06 bandgap. PBE bandgap predictions are also reported for comparison.

| Refcode | Predicted HSE06 | Predicted PBE | Conductivity (S cm$^{-1}$) |
|---|---|---|---|

| | (eV) | (eV) | |
|---|---|---|---|
| BITCUE[86] | 0.117 | 0.205 | $2.5 \times 10^{-1}$ |
| ESOSOV[87] | 0.488 | 0.116 | $5.4 \times 10^{-1}$ |
| ECITOD[88] | 0.643 | 0.000 | $1.9 \times 10^{-3}$ |
| FUVZEE[89] | 1.056 | 0.000 | $2.0 \times 10^{-3}$ |
| FUVZEE01[89] | 1.103 | 0.000 | $2.0 \times 10^{-3}$ |
| FAFJAZ[90] | 1.107 | 0.000 | $1.2 \times 10^{-1}$ |
| GEFKAG[91] | 1.141 | 0.327 | $1.4 \times 10^{-2}$ |
| MAKPOH[92] | 1.147 | 0.126 | $5.0 \times 10^{-6}$ |
| BALLOP[93] | 1.220 | 0.000 | $1.3 \times 10^{-5}$ |
| VUSNUT[90] | 1.225 | 0.000 | $2.7 \times 10^{-1}$ |
| FUVZUU[89] | 1.306 | 0.000 | 9.0 |
| KUCHAS[94] | 1.331 | 0.474 | $4.0 \times 10^{-5}$ |
| DEYWAI[95] | 1.392 | 0.315 | $3.0 \times 10^{-5}$ |
| DIFWEY01[96] | 1.399 | 0.142 | $2.5 \times 10^{-6}$ |
| KEDHIN[97] | 1.400 | 0.939 | $5.0 \times 10^{-4}$ |
| MAKPUN[92] | 1.412 | 0.116 | $1.0 \times 10^{-6}$ |
| ESOSUB[87] | 1.485 | 0.550 | $2.4 \times 10^{-4}$ |
| LUHFOM[98] | 1.553 | 0.000 | $4.4 \times 10^{-5}$ |

A representative re-discovered structure is GEFKAG[91], whose systematic chemical name is "catena-[tris((bis($\eta^5$-cyclopentadienyl)-cobalt(iii)) dimethylammonium) tris($\mu$-3,6-dichlorobenzene-1,2,4,5-tetrolato trianion radical)-di-iron(iii) N,N-dimethylformamide solvate]". The model assigns this framework a predicted HSE06 bandgap of 1.141 eV (PBE: 0.327 eV), placing it within the top conductive MOFs identified (reported conductivity: $1.4 \times 10^{-2}$ S cm$^{-1}$ to $5.1 \times 10^{-4}$ S cm$^{-1}$). The phrase "tetrolato trianion radical" identifies a benzene ring bearing four deprotonated oxygen donors coordinated to metal centers, while the "radical" designation states a partially occupied ligand-centered $\pi$ orbital. Such ligand-based radical states introduce partially filled $\pi$ levels close to the Fermi level, reducing the energetic separation between occupied and unoccupied ligand orbitals. The presence of "di-iron(III)" further indicates open-shell transition-metal nodes with accessible d orbitals that can electronically couple with the ligand $\pi$ system through metal-ligand orbital interactions. Together, these features describe a framework in which radical-bearing aromatic ligands interact with open-shell metal centers, enabling delocalization between ligand $\pi$ orbitals and metal d orbitals across the extended structure. Because these electronically active motifs are explicitly encoded in the systematic chemical name, the model is able to associate such descriptors with narrow bandgaps.

The Tl(TCNQ) MOF provides a stringent test of whether systematic nomenclature can resolve subtle connectivity differences between polymorphs. The structure's phase I (ESOSUB[87]), catena-[($\mu_5$-7,7,8,8-tetracyanoquinodimethane radical anion-N,N,N',N',N'')-thallium(i)], was reported to convert in the solid state to Phase II (ESOSOV[87]), catena-[($\mu_4$-7,7,8,8-tetracyanoquinodimethane radical anion-N,N,N',N')-thallium(i)], upon exposure to ambient water vapour[87]. The detailed structural comparison is shown in Figure S5-9 in Supplementary Information. Both systematic names contain the same ligand fragment, "Tetracyanoquinodimethane radical anion" (TCNQ$^-$), coordinated to thallium(I), indicating that the two polymorphs share the same electronically active ligand and metal center. Experimentally, however, the two phases exhibit markedly different charge-transport behavior: Phase I shows alternating separations between adjacent radical TCNQ units, inducing partial dimerization and conductivity of $2.4 \times 10^{-4}$ S cm$^{-1}$ , whereas Phase II adopts a more uniform $\pi$-stacking arrangement and exhibits a significantly higher conductivity of $5.4 \times 10^{-1}$ S cm$^{-1}$. The structural distinction between the two polymorphs is reflected in the connectivity descriptors encoded in the systematic names. In ESOSUB, the $\mu_5$ coordination mode and five nitrile donor designations (N,N,N',N',N'') indicate that the TCNQ radical anion connects to five thallium centers, whereas in ESOSOV the $\mu_4$ descriptor and four nitrile donors (N,N,N',N') correspond to a reduced coordination. Our model predicts an HSE06 bandgap of 1.485 eV for ESOSUB (PBE: 0.550 eV), while ESOSOV is assigned a substantially smaller HSE06 bandgap of 0.488 eV (PBE: 0.116 eV), consistent with the more electronically coherent stacking observed experimentally. This example illustrates that systematic chemical names encode connectivity information that the model can leverage to distinguish closely related structures and infer differences in their electronic properties.

As summarized in Table 3, among the remaining 32 structures that have not been previously reported as electrically conductive, we identified a subset of 10 promising candidates with the lowest predicted bandgaps.

**Table 3.** Promising conductive MOFs identified via screening of 105,328 previously unseen structures from the Cambridge Structural Database (CSD) MOF subset.

| Refcode | Predicted HSE06 (eV) | Predicted PBE (eV) | Systematic chemical name |
|---|---|---|---|
| ALUTUX[99] | 1.196 | 0.337 | catena-(bis($\mu_2$-4',5'-bis(methylsulfanyl)-1,1',3,3'-tetrathiafulvalene-4,5-dicarboxylato)-bis($\mu_2$-4,4'-bipyridine)-diaqua-di-manganese(ii) acetonitrile solvate) |
| DECWUE[100] | 1.341 | 0.289 | catena-(($\mu_2$-4,4'-bipyridine)-bis($\mu_2$-dimethyldithiophosphonato)-di-copper(i)) |
| LEXSAL[101] | 1.354 | 0.178 | catena-[bis($\mu$-2-{[(2-oxidophenyl)methylidene]amino}propanoato)-bis(2,2'-bipyridine)-manganese(ii)-manganese(iii) perchlorate hemihydrate] |
| LURFAF[102] | 0.849 | 0.138 | catena-(bis($\mu_2$-$\eta^2$,$\eta^2$-syn-benzoquinone)-tetrakis($\mu_2$-acetato-o,o')-bis($\mu_2$-4,4'-bipyridine)-tetra-copper(i) hydroquinone solvate) |
| OCOKUP[103] | 1.001 | 0.855 | catena-(tris($\mu$-4,4'-bipyridine)-hexakis($\mu$-cyano)-dicyano-diethanol-tetra-manganese bis(4,4'-bipyridine)) |
| QAGTIB05[103] | 1.048 | 0.000 | catena-[bis($\mu$-iodo)-bis($\mu$-4,4'-bipyridine)-di-copper(i)] |
| SEZDAG[104] | 1.332 | 0.513 | catena-(bis($\mu$-chloro)-($\mu$-2,7-bis(pyridin-3-yl)benzo[lmn][3,8]phenanthroline-1,3,6,8(2H,7H)-tetrone)-di-copper) |
| UJUJEP[105] | 1.390 | 0.000 | catena-(($\mu_6$-1,1',3,3'-tetrathiafulvalene-4,4',5,5'-tetracarboxylato)-bis(2,2'-bipyridine)-di-manganese(ii) dihydrate) |
| YIDSAF07[106] | 1.381 | 0.510 | catena-[bis($\mu$-chloro)-bis($\mu$-4,4'-bipyridine)-di-copper(i)] |
| YUNHIB[107] | 1.278 | 0.768 | catena-[bis($\mu$-cyano)-bis($\mu$-2,6-bis(pyrazin-2-yl)-4,4'-bipyridine)-di-copper(i) carbon dioxide] |

## 2.5   Language-model reasoning with systematic MOF nomenclature

Inspired by the pioneering work of Yaghi and collaborators on the development of the RetChemQA dataset[108] for machine learning applications, we explored how chemical nomenclature influences the reasoning capabilities of large language models (LLMs). RetChemQA provides a curated benchmark of MOF-related question–answer pairs, but primarily relies on shorthand framework identifiers (e.g., "MOF-14") that obscure explicit chemical detail.

Building on this foundation, we investigate whether replacing such identifiers with systematic IUPAC-style nomenclature alters the quality and interpretability of model reasoning.

This experiment builds directly on the earlier embedding-based analyses, where systematic MOF names were shown to encode chemically meaningful information for downstream prediction tasks. Here, we examine whether the same information content, when presented in linguistic form, can be leveraged by large language models to support higher-level chemical reasoning and interpretability. To ensure a controlled comparison, the raw systematic names are used directly as text input without any chemistry-specific preprocessing or feature engineering. Their preserved symbolic structure is therefore processed solely through standard tokenization, allowing the model to learn chemical meaning directly from nomenclature itself rather than from externally imposed chemical rules.

To isolate the effect of naming conventions, we construct two parallel versions of a RetChemQA-derived question-answer dataset. In the first version, MOFs are referred to using the original shorthand identifiers. In the second version, these identifiers are systematically replaced with their corresponding IUPAC-style names, for example "*catena*-(bis($\mu_6$-(benzene-1,3,5-triyl)-4,4',4''-tribenzoato)-triaqua-tri-copper(ii))". All other aspects of the dataset, including question types, answers, and supervision signals, are held fixed. Identical LLM architectures (Llama-3.2-3B-Instruct[109]) are fine-tuned on each dataset under the same training protocol, enabling a controlled comparison of how naming conventions alone affect model reasoning behavior.

Rather than focusing exclusively on task-level accuracy, we analyze how chemically explicit naming affects both the relevance and interpretability of model outputs. As illustrated in Figure 8, models trained on systematic names consistently produce more chemically coherent and complete answers across multiple question types compared to models trained on shorthand identifiers.

Figure 8a presents an example involving chemical-formula inference. When provided with a systematic name, the model correctly reconstructs the framework composition by leveraging explicit lexical cues embedded in the nomenclature. Terms such as "tri-copper(ii)," "tribenzoato," and "triaqua" encode metal stoichiometry, linker identity, and coordinated ligands, respectively, enabling recovery of the correct chemical formula. In contrast, when only a shorthand label is supplied, no chemically informative input units are available; SHAP attributions become diffuse, and the predicted formula deviates from the crystallographic record. This example illustrates typical reasoning behavior rather than exhaustive performance characterization. A similar pattern is observed for synthesis-related queries (Figure 8b). Systematic names provide explicit cues for inferring precursor identities and reaction components. SHAP attributions concentrate on chemically meaningful input units, such as metal oxidation states and linker descriptors, each contributing positively to the generated answer. By contrast, shorthand identifiers lack such semantic structure, leading to incomplete or misleading predictions and poorly localized attributions.

**a**

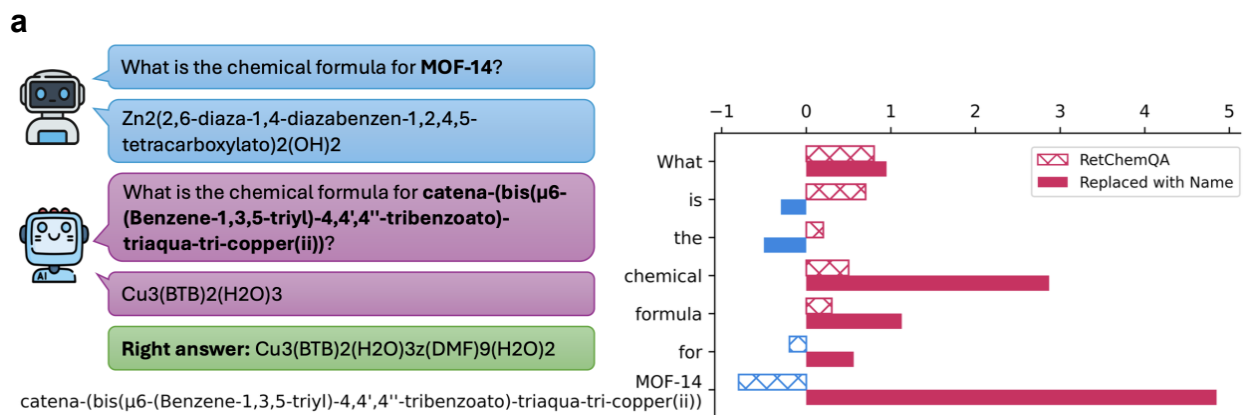

**b**

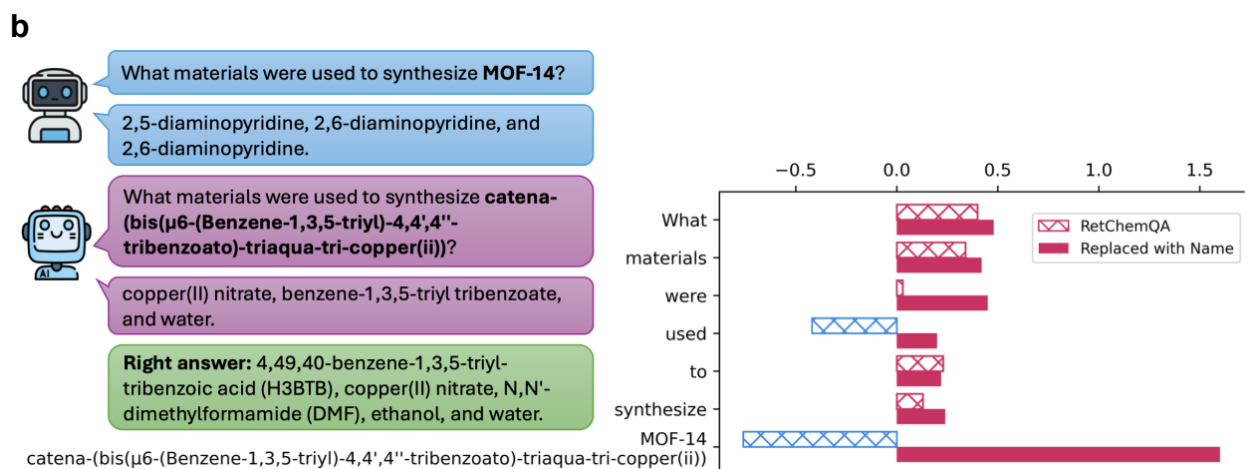

catena-(bis(μ6-(Benzene-1,3,5-triyl)-4,4',4''-tribenzoato)-triaqua-tri-copper(ii))

**Figure 8. Impact of systematic chemical names on language model reasoning and interpretability.** The blue boxes represent model answers generated using the original RetChemQA dataset with simplified identifiers (e.g., "MOF-14"), the **red boxes** represent answers generated using the modified dataset with systematic IUPAC-style names, and the **green boxes** indicate the reference ground-truth answers. SHAP values shown in the right panels quantify the contribution of each input to the prediction: **positive values** indicate units that improve model accuracy by contributing chemically meaningful information, while **negative values** indicate units that reduce accuracy or mislead the model. **(a)** Example responses to a chemical formula query, showing that systematic names lead to chemically more consistent answers than simplified identifiers. **(b)** Example responses to a synthesis-related query, where systematic names improve both the completeness of the generated answer and the interpretability of unit contributions.

Across question types, SHAP analysis consistently reveals that input units derived from systematic nomenclature contribute positively to model predictions, whereas shorthand identifiers provide little interpretable signal. These results support two complementary conclusions. First, systematic chemical names act as effective semantic anchors, embedding explicit chemical entities that enable LLMs to disambiguate queries and access domain-relevant knowledge. Second, the resulting reasoning process is more interpretable, with attribution patterns aligned to chemically meaningful components of the input.

Taken together, this analysis demonstrates that the information encoded in systematic MOF nomenclature is not limited to embedding-based prediction tasks, but can also be leveraged by generative language models to support chemically grounded reasoning. These findings suggest that enriching identifier-based corpora with standardized chemical names can improve both the fidelity and interpretability of LLM-based workflows in materials informatics. Additional computational details, extended evaluations, and mixed-metal case studies are provided in Supplementary Information S6.

# 3    Conclusions

In summary, we demonstrate that systematic chemical names offer a powerful and structure-independent representation for machine learning tasks involving MOFs. By leveraging pretrained language models, ReadMOF converts IUPAC-style nomenclature from the Cambridge Crystallographic Database (CSD) into continuous embeddings that implicitly reflect molecular architecture and coordination patterns, all inferred from a linguistic structure. Despite relying solely on nomenclature, the resulting embeddings capture global structural similarity, enable chemically meaningful clustering, and support high-fidelity retrieval across large MOF datasets. Building on this foundation, ReadMOF achieves accurate prediction of both structural and electronic properties and further enhances generative reasoning when coupled with large language models (LLMs). These results highlight systematic nomenclature as a lightweight, interpretable, and generalizable alternative to coordinate-based representations, offering a powerful route toward accelerated discovery and rational design of reticular materials.

# 4    Acknowledgement

P.Z.M. acknowledges support from the Department of Science, Innovation and Technology (DSIT) and the Royal Academy of Engineering under the Industrial Fellowships program (IF2223-110). He also acknowledges support by UK government through Innovate UK (Grant 10098491). B.M. is supported by the Polish National Agency for Academic Exchange (decision no. BPN/BEK/2024/1/00218/DEC/1). A.M.C acknowledges the Royal Commission for the Exhibition of 1851 for a Research Fellowship.

# TOC

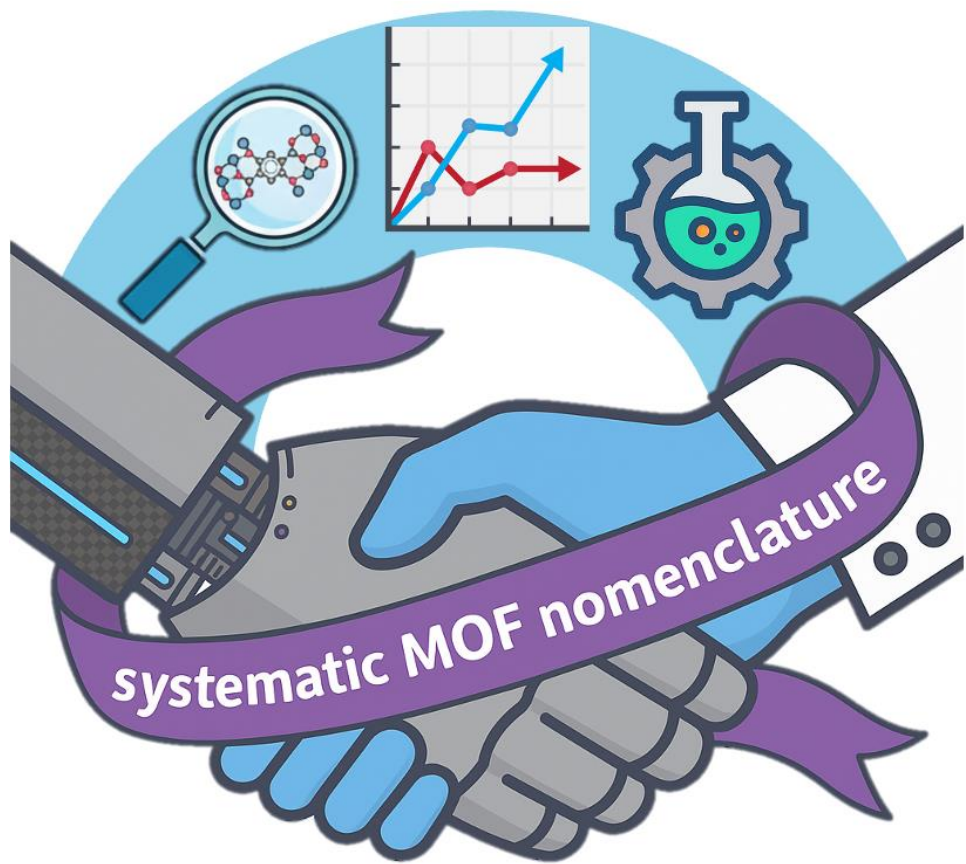